\documentclass[preprint,12pt]{elsarticle}

\usepackage{amssymb}
\usepackage{graphicx} % for pdf, bitmapped graphics files
\usepackage{epsfig} % for postscript graphics files
\usepackage{mathptmx} % assumes new font selection scheme installed
\usepackage{amsmath} % assumes amsmath package installed
\usepackage{xcolor}
\usepackage{url}
\usepackage{soul}
\usepackage{mathrsfs}
\usepackage{subcaption}
\usepackage{float}
\usepackage[11pt]{moresize}
\usepackage[english]{babel}
\usepackage[utf8]{inputenc}
\usepackage{csquotes}
\usepackage{makecell}
\usepackage{amsthm}
\newtheorem{theorem}{Theorem}
\newtheorem{remark}{Remark}
\newtheorem{definition}{Definition}
\usepackage{lipsum} 
\journal{Arxiv}

\begin{document}

\begin{frontmatter}



\title{Aerial Transportation Control of Suspended Payloads with Multiple Agents\corref{funding}}


\cortext[funding]{This work was supported by the Mexican National Council of Science and Technology CONACYT.} 

\author[fati]{Fatima Oliva-Palomo}

\affiliation[fati]{organization={Center for Research in Mathematics CIMAT AC, campus Zacatecas},%Department and Organization
            city={Zacatecas},
            postcode={98160}, 
            state={Zacatecas},
            country={Mexico}}
            
 \author[die]{Diego Mercado-Mercado\corref{cor1}}

\affiliation[die]{organization={Investigadores CONACYT at Center for Research in Mathematics CIMAT AC, campus Zacatecas},%Department and Organization
            addressline={Calle Lasec y Andador Galileo Galilei, Manzana 3, Lote 7 Quantum Ciudad del Conocimiento}, 
            city={Zacatecas},
            postcode={98160}, 
            state={Zacatecas},
            country={Mexico}
            }    
\cortext[cor1]{corresponding author.}  
\ead{diego.mercado@cimat.mx}

\author[pedro]{Pedro Castillo}

\affiliation[pedro]{organization={Sorbonne Universitès, Université de Technologie de Compiègne, CNRS, Heudiasyc UMR 7253},%Department and Organization
            city={Compiègne},
            postcode={60200}, 
            country={France}}
    
\begin{abstract}
In this paper we address the control problem of aerial cable suspended load transportation, using multiple Unmanned Aerial Vehicles (UAVs). First, the dynamical model of the coupled system is obtained using the Newton-Euler formalism, for $n$ UAVs transporting a load, where the cables are supposed to be rigid and mass-less. The control problem is stated as a trajectory tracking directly on the load. To do so, a hierarchical control scheme is proposed based on the attractive ellipsoid method, where a virtual controller is calculated for tracking the position of the load, with this, the desired position for each vehicle along with their desired cable tensions are estimated, and used to compute the virtual controller for the position of each vehicle. This results in an underdetermined system, where an infinite number of drones' configurations comply with the desired load position, thus additional constrains can be imposed to obtain an unique solution. Furthermore, this information is used to compute the attitude reference for the vehicles, which are feed to a quaternion based attitude control. The stability analysis, using an energy-like function, demonstrated the practical stability of the system, it is that all the error signals are attracted and contained in an invariant set. Hence, the proposed scheme assures that, given well posed initial conditions, the closed-loop system guarantees the trajectory tracking of the desired position on the load with bounded errors. The proposed control strategy was evaluated in numerical simulations for three agents following a smooth desired trajectory on the load, showing good performance.

\end{abstract}


% \begin{highlights}
% \end{highlights}

\begin{keyword}
Suspended Payload Control \sep UAVs \sep Multi-Agent Systems \sep Aerial Transportation \sep Attractive Ellipsoid Method \sep Practical Stability
\end{keyword}

\end{frontmatter}



\section{Introduction}
\label{sec:intro}
%- Justificacion

Small scale Unmanned Aerial Vehicles (UAVs) have received increased attention in the last decades, thanks to their great potential in civilian applications such as surveillance, monitoring, photography, structures inspection, fast deployment of food and medical equipment, etc. \cite{KumarSurvey}. A clear example of this, is the use of UAVs to transport packages quickly, specially in urban environments \cite{Ollero2022}, optimizing the distance and time required by avoiding and alleviating traffic congestion. Such potential has not passed unnoticed, to the point that the most important transnational delivery companies have shown great interest, and invested time and money to develop aerial transportation systems using drones capable to deliver packages quickly.

\begin{figure}[t]
    \centerline{\includegraphics[width=0.98\textwidth]{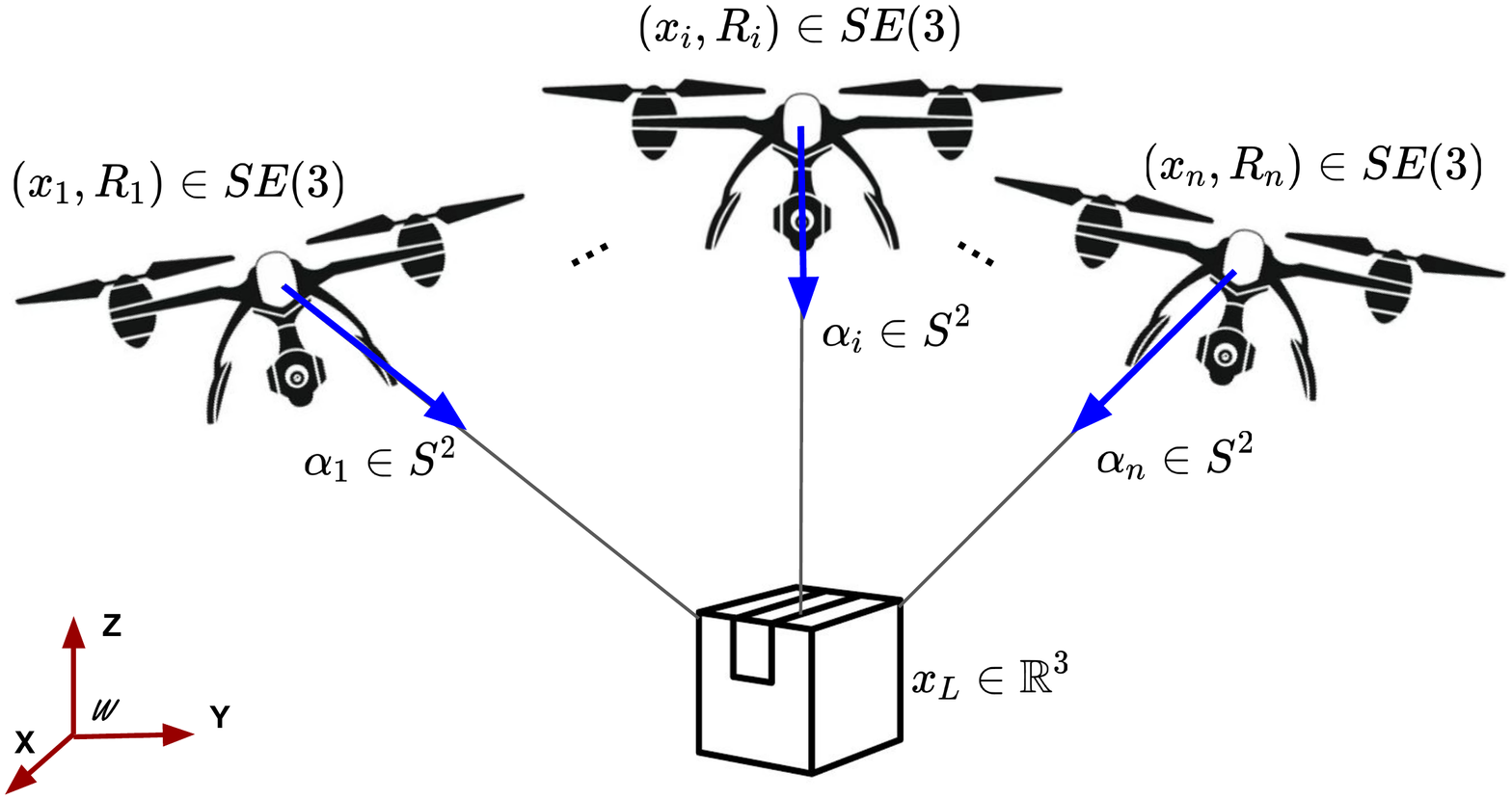}}
    \caption{Aerial transportation system with multiple UAVs. The cables are assumed rigid and mass-less, while the payload is considered as a punctual mass carried by $n$ aerial vehicles.}
    \label{fig:aerialTrasnport}
\end{figure}

Aerial transport using UAVs presents important design challenges, given that drones normally have reduced payload capabilities, but it also appears as an interesting control problem, provided that the payload mass and inertia matrix may be unknown to the system, also, it may be required to optimize the energy employed in the mission, since the autonomy of the system is considerably reduced with larger payloads, moreover, the payload itself introduces uncertainty and disturbances to the system, for instance, the center of mass of the system is shifted once a payload is added. More in particular, we center our attention to the problem o cable suspended payloads, a problem that arises from missions where the UAV hooks the payload on flight, without the need of landing, a useful feature which allows to avoid the risk of landing in hazardous terrains and situations. Cable suspended aerial transportation (see Fig. \ref{fig:aerialTrasnport}) has attracted recently the attention from the research community, specially from the automatic control perspective, since the resulting system has more degrees of under-actuation, and the suspended payload introduces oscillations and disturbances \cite{Guerrero2021}. 

Accordingly, a few works are to be found in the literature \cite{VillaICUAS2018}, where the former studies were centered in the study of a single vehicle under idealized conditions, considering a punctual mass suspended by a rigid mass-less cable \cite{Guerrero2015, Guerrero2017}. From there, some works have popped up considering different variations to the problem, assuming conditions closer to a real application, such as assuming the load as a rigid rod \cite{Villa2021, Goodman2022-ue}, or a rigid body \cite{Lee2014,Wu2014}, or considering that the cables possess elasticity \cite{Goodman2022-ue,Kotaru2017} or are flexible \cite{Goodarzi2015}, or that more than one vehicle contributes to carry the load \cite{Bernard2011, Fink2011, Lee-Kumar2013, Pereira2017}. 

Nevertheless, small aerial vehicles, particularly multirotor rotor-crafts, suffer from their flight time autonomy and their payload capacity, while increasing their size is undesirable for safety reasons, since the risk of severe accidents scales up with size, to the point of endangering human lives. In that sense, aerial transportation using multiple drones appears as an appealing and challenging alternative, with great potential for applications in the real world, allowing to distribute the payload weight among a team of small UAVs. Furthermore, such multi-agents scheme adds redundancy and enables fault tolerance capabilities to the system, such that the mission can be accomplished even in the case of a problem with some of the agents.

\subsection{Related Works}
\label{sec:related}
Besides its applicability, the problem of aerial transportation with multiple drones and cable suspended loads turns out to be a very interesting and challenging problem from the automatic control point of view, given that it deals with a nonlinear system of high order, with multiple degrees of freedom, and an important number of degrees of under-actuation, not to mention the parametric uncertainties and disturbances induced by the other agents or the payload itself, or even external phenomena affecting the system in real applications, such as wind gusts. 

Only a handful of works have studied the problem of aerial transportation of cable suspended payloads with multiple agents \cite{Ollero2022}. For instance, in \cite{Carino2020}, a passivity based control is proposed for two UAVs, but only in the $XZ$ plane. The proposed control strategy was validated only in numerical simulations. Also, the authors in \cite{Villa2021} present an adaptive dynamic compensator, for the case of two UAVs carrying a cable suspended rigid rod. The cables are considered rigid, and experimental validation is provided. Moreover, in \cite{Gassner2017}, the authors propose a vision-based cable suspended load transport with two quadrotors, in a leader follower approach. The load is a rigid rod, but it is modeled as a point mass. An LQR (Least Quadratic Regulator) controller is used. The performance of the transportation system is validate in real experiments. Nevertheless, these works only consider a fixed small number of agents, ranging from two to four.

On the other hand, there are some works that study the more general transportation problem for $n\in \mathbb{Z}^+ \mid n>2$ UAVs \cite{Michael2011-lc,Jackson2020,Zeng2020}, but they focus mainly on the motion planning problem, using simple open-loop controllers, it is, without considering any feedback on the pose of the load. This is somehow similar to the flight formation problem, with the load acting as an external disturbance. Such is the case of the work at \cite{Sreenath2013}, where the dynamic model of the transportation system is presented using the hybrid systems formalism, where the payload is modeled either as a point mass or as a rigid body, and the cables are considered rigid and mass-less, with strictly positive tension. The hybrid systems are used to model the cases when an agent stops contributing with the payload transportation, i.e., when for some reason its cable tension equals zero. Then, a differential flatness approach is employed to design a control strategy. However, such controller lacks of feedback in the payload, i.e. it acts in open-loop with respect to the payload. Experimental results with $n=3$ UAVs validate the proposed scheme.
%% Motion planning, only open loop control
%\cite{Michael2011-lc}, \cite{Jackson2020}, \cite{Zeng2020}
%% Kumar : hybrid system, differential flatness, rigid cable, rigid body, n-agents, experiments, but only open loop control

Probably the most studied controller schemes in the literature for the cable suspended payload transportation control problem with $n$ UAVs is the use of Geometric controllers that work directly in the Special Euclidean group in three dimensions $SE(3)$, considering the payload as a rigid body, along with rigid and mass-less cables \cite{Wu2014,Lee2014,Lee2018}. Also, in 
\cite{Goodarzi2015} a geometric control is proposed, but the cables are assumed as flexible instead. The Euler-Lagrange formalism is normally used to obtain the dynamic model. These works propose interesting solutions to the control problem under consideration, with formal stability analysis, nonetheless, such controllers are not straightforward to implement in real-time experiments, hence their validation remains only in the simulation stage.
%% Geometric control, rigid body, simulation, n-agents
%\cite{Wu2014}, \cite{Lee2014}, \cite{Lee2018}
%% flexible cables
%\cite{Goodarzi2015}

%%Flying parallel robot
Another interesting recent related work is the one presented in \cite{Six2021}, where a transportation system with multiple UAVs is presented, but instead of a cable suspended payload, multi-links legs are used to hold a rigid payload up, something like a flying parallel robot. Experimental results demonstrate the feasibility of the system.

\subsection{Contributions}
In this work, we deal with the control problem of load trajectory tracking in an aerial transportation system conformed by $n\leq 2$ UAVs and a cable suspended payload, considered as a punctual mass, and rigid mass-less cables. In contrast to former multi-agent aerial systems such as formation flight or consensus based, here the control problem is defined directly on the load position. The proposed closed-loop hierarchical controller is designed using the attractive ellipsoid method, where the resultant of the cable tensions is used as a virtual controller to assure trajectory tracking on the load, then, the desired tension on each UAV cable is obtained from the virtual load controller, resulting in an under-determined system, from where additional constrains can be imposed to obtain an unique solution. Afterwards, the desired UAV position is feed to a position controller for each agent, using its desired attitude as a virtual controller. Finally, a quaternion based controller is employed to command the attitude of each drone. 

The stability of the closed-loop system was analyzed using the attractive ellipsoid method, guaranteeing the practical stability of the system as long as the initial conditions are well posed, and the desired load trajectories are smooth, with bounded relative accelerations between the agent and the load. In other words, the load trajectory tracking is assured with bounded errors, which depend mainly on the disturbances caused by the virtual controllers transient response. Numerical simulations validate the good performance of the proposed strategy.

Focusing in the general problem of cable suspended payload aerial transportation with $n \leq 2$ agents, in this paper we proposed a closed-loop controller, considering feedback from the load position, which is the main goal in the control formulation, this is in contrast with other works on the literature which focus mainly in the motion planning problem and propose open loop solutions without feedback from the load \cite{Michael2011-lc,Jackson2020,Zeng2020}. This work also offers an alternative solution to the geometric control approach proposed in the works \cite{Wu2014,Lee2014,Lee2018, Goodarzi2015}, in a more intuitive framework. 

Moreover, the use of the attractive ellipsoid method allows to study the robustness of the system by providing the size of the stability region using the attractive matrix. Furthermore, the method defines a procedure to obtain the gains to assure the smallest invariant set contemplating the bounds of the disturbances of the coupled system. Additionally, taking into consideration the disturbances due to the virtual controllers, an analysis of the necessary conditions for stability is discussed.  

%The proposed scheme is closed-loop continuous robust? control for $n$ agents, that assures bounded errors in the position of the load.

The contribution of the present work can be summarized as follows:

\begin{itemize}
    \item A hierarchical controller based in the attractive ellipsoid method is proposed for the load trajectory tracking problem in cable suspended payload aerial transportation systems with $n$ agents. The proposed closed-loop system is continuous and assures practical stability.
    \item The stability analysis was carried out using a Lyapunov-like function and considering the disturbances due to the virtual controllers. Also, the operating conditions were discussed, including the initial conditions and the relative accelerations between the load and the agents. Numerical simulations validated the obtained results.
 %   \item The attitude controller is singularity free, given that quaternions representation is used.
    %control robusto para n agentes, basado en quaterniones, basado en elipsoide invariante
    %analisis de estabilidad y discusion sobre las condiciones de operacion
\end{itemize}

The reminder of the paper is organized as follows. In Section \ref{sec:problem} the control objective is established and the dynamic model of the aerial transportation system with multiple agents is presented, while. Then, in Section \ref{sec:control} we introduce the control strategy, and analyze the closed-loop stability by means of an energy function. Thereafter, in Section \ref{sec:simulation}, the performance of the closed-loop system is studied in numerical simulations, showing good results. Finally, in Section \ref{sec:conclusions}, some conclusions and future works are drawn.

\section{Problem Statement}
\label{sec:problem}

\subsection{Dynamic model}
\label{sec:model}
Let us consider an aerial transportation system composed of $n \in \mathbb{Z}^+ $ Unmanned Aerial Vehicles (UAVs) carrying a cable suspended payload, under the following assumptions
\begin{itemize}
    \item The load can be considered as a punctual mass.
    \item All the cables are mass-less.
    \item The cables are rigid with positive tension.
    \item The aerodynamic effects are not considered.
    \item The air friction is negligible.
    \item The desired trajectories are smooth, with bounded accelerations.
    %\item External disturbances such as wind act as additive forces, unknown but bounded. 
%\textcolor{blue}{XXXXXXXX}
\end{itemize}
Each of the agents (UAVs) has a mass $m_i$, and an inertial matrix $J_i$, then, the dynamics of motion for the $i$-th agent can be modeled as
\begin{equation}
    m_i\ddot x_i= f_iR_i(\bar{q}_i)e_3-m_ige_3+T_i\alpha_i
    \label{model:pos}
\end{equation}
\begin{equation}
    \dot {\bar{q}}_i= \frac{1}{2} \bar{q}_i\otimes \bar{\Omega }_i
    \label{model:att}
\end{equation}
\begin{equation}
    J_i\dot \Omega_i+\Omega_i\times J_i\Omega_i=\tau_i 
    \label{model:ori}
\end{equation}
where $x_i\in \mathbb{R}^3$ stands for the position of the $i$-th UAV's center of mass, with respect to an inertial reference frame $\mathbf{\textit{W}}$, as depicted in Fig.\ref{fig:aerialTrasnport}, $R_i\in SO(3)$ is a rotation matrix providing the attitude of each UAV. $f_i\in \mathbb{R}$ represents the total thrust force produced by the rotors, $e_3 \triangleq [0\ 0\ 1]^T$, $g$ is the gravity constant. $T_i\in \mathbb{R}$ are the cable tensions, and $\alpha_i\in \mathbb{S}^2$ is a unit vector representing the orientation of each cable connecting the $i$-th UAV with the load. Moreover, $\Omega_i\in \mathbb{R}^3$ is the angular velocity in the body fixed frame $\mathbf{\textit{B}}_i$, and $\tau_i\in \mathbb{R}^3$ are the input torques, 
$\bar{\Omega}_i=[0,\Omega_i^{\mathrm{T}}]^{\mathrm{T}}$ is a pure quaternion of the vector $\Omega_i$. The attitude is parameterized by the unit quaternion \cite{Trawny} $\bar{q}_i = [q_{i0},q_i ^{\mathrm{T}}]^{\mathrm{T}} \in \mathbb{Q}$, where $q_{i0} \in \mathbb{R}$ and $q_i =[q_{i1},q_{i2},q_{i3}]^\mathrm{T}\in \mathbb{R}^3$, while its inverse quaternion is given by $\bar{q}_i^* = [q_{i0},- q_i ^{\mathrm{T}}]^{\mathrm{T}}$.
The quaternion product $\otimes$, between two unit quaternions $p,q\in \mathbb{Q}$ is given by \cite{mapeo2021}
\begin{equation}
\overline{p} \otimes \overline{q} =
\begin{bmatrix}
{p_0}&  {-{p}^\mathsf{T}}   \\    {p} & { I{p_0} + [{p} \times ]} 
\end{bmatrix} \begin{bmatrix}
q_0 \\ q 
\end{bmatrix}
\label{q-product}
\end{equation} 
where $p\times$ maps the vector $p\in \mathbb{R} ^3$ to a skew symmetric matrix $so(3)$, and the identity matrix is $I\in \mathbb{R}^{3\times 3}$. Furthermore, the rotation matrices $R_i$ are parameterized with unit quaternions as follows

\begin{equation}
R_i(\overline{q})=\small \begin{bmatrix}
{1-2q_{i2}^2-2q_{i3}^2}&{2q_{i1}q_{i2}-2q_{i0}q_{i3}}&{2q_{i1}q_{i3}+2q_{i0}q_{i2}}\\
{2q_{i1}q_{i2}+2q_{i0}q_{i3}}&{1-2q_{i1}^2-2q_{i3}^2}&{2q_{i2}q_{i3}-2q_{i0}q_{i1}}\\
{2q_{i1}q_{i3}-2q_{i0}q_{i2}}&{2q_{i2}q_{i3}+2q_{i0}q_{i1}}&{1-2q_{i1}^2-2q_{i2}^2}\\
\end{bmatrix}.
\label{model:RotMat}
\end{equation}

The cables are assumed to be mass-less and rigid, with length $L_i$, hence, the payload is subject to the following restriction with respect to each aerial drone

\begin{equation}
    x_i=x_L-L_i\alpha_i
    \label{model:restriction}
\end{equation}

The payload is considered as a punctual mass with mass $m_L$, with position $x_L\in \mathbb{R}^3$, then, the dynamics of the load are subject to the following equation

\begin{equation}
    m_L\ddot x_L= -m_Lge_3-\Sigma T_i\alpha_i
    \label{model:load}
\end{equation}

\subsection{Control problem}

%- XXXXXX Control de transporte de carga con multiples agentes
We consider the trajectory tracking problem on the load, where the main goal is to design a closed-loop controller that assures that the position of the load $x_L(t)$, and its time derivatives, tend to a desired time varying reference $x_{Ld}(t)$, it is
$\{x_L, \dot x_L \} \rightarrow \{x_{Ld},\dot x_{Ld} \}$, considering the dynamic of the system  \eqref{model:load}, subject to the constrains \eqref{model:restriction} and the coupled dynamics of \eqref{model:pos}, guaranteeing that the error trajectories $\{x_e, \dot x_e\}$ are confined to an attractive invariant set. The controller is continuous and the algorithm considers the underactuation of the aerial vehicles and the coupled dynamics of the other drones.

\begin{remark}
Note that the control objective is defined on the payload's position, not on the UAVs. In contrast with formation flight controllers \cite{MAHMOOD2017852,Mercado2013}, here the formation of the agents is irrelevant as long as they respect the system restrictions imposed by the cables \eqref{model:restriction}, accordingly, there exist infinity configurations (formations) where the agents are able to transport the load to its desired reference trajectory. Additional constrains can be imposed to the system in order to obtain a unique solution.
\end{remark}

\section{Control Strategy}
\label{sec:control}
\begin{figure}[t]
\vspace{-1cm}
    \centerline{\includegraphics[width=0.98\textwidth]{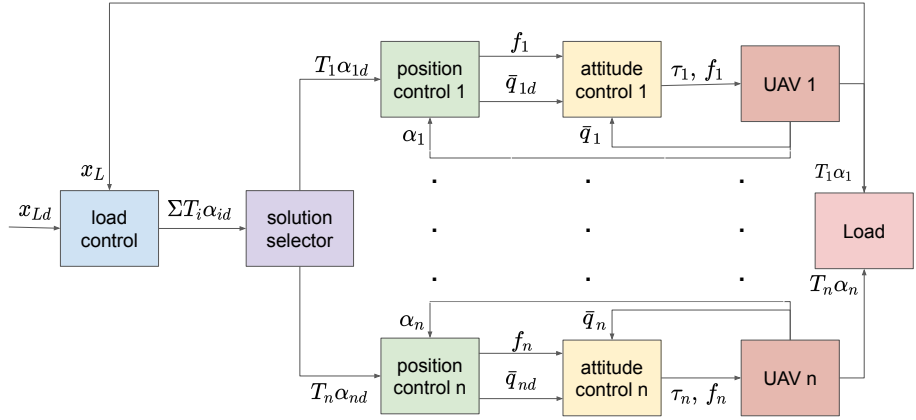}}
    \vspace{-2cm}
    \caption{Blocks diagram of the closed-loop system. A hierarchical control scheme is proposed, where the load control uses the resultant tensions as a virtual control input, then the target tensions are distributed to the agents and a position controller computes the desired thrust force vector in order to guarantee that each agent moves to its desired position. Finally, an attitude controller extracts the desired orientation of each drone and computes the required input torque. }
    \label{fig:block}
\end{figure}
% control strategy abstract
As presented in Fig. \ref{fig:block}, where the block diagram of the control strategy is depicted, a hierarchical controller is proposed, where a virtual controller is employed to control the load $u_L$, which represents the desired resultant tension, it is the sum of the desired tensions in each cable $\sum_{i=1}^{n}  \left( T_{id} \alpha_{id} \right) $. From the Load's control law $u_L$, the desired resultant force on the load is distributed among the vehicles and the desired position of each agent is computed using the desired direction of the cable $\alpha_{id}$ and the desired tension $T_{id}$. Note that the resultant system is under-determined, and there exist an infinite number of valid solutions, which means that there exists an infinite number of valid configurations for the agents to get the load to its desired position, however, it is enough to impose additional constrains on the desired tensions to obtain only one valid solution. Then, a virtual controller for the position of the vehicles is computed, providing the desired input force for each quadrotor, this results in a desired thrust force $f_{id}$ and a desired attitude quaternion $\bar{q}_{id}$, which is parsed to the attitude control for each vehicle. Therefore, assuring the desired attitude is reached fast enough $\bar{q}_{i} \rightarrow \bar{q}_{id}$, implies that each of the vehicles tends to their desired position $x_i \rightarrow x_{id} $, and the tension in the cables tends to the desired tension, resulting in the tracking of the desired position by the load, it is $x_L \rightarrow x_{Ld} $. 

In this section, first the attractive ellipsoid method is described, afterwards, the attitude and virtual position controller are obtained, then, the position and load control laws, used in the virtual controllers, are developed.

% A hierarchical control scheme is proposed, where the load control uses the resultant tensions as a virtual control input, then the target tensions are distributed to the agents and a position controller computes the desired thrust force vector in order to guarantee that each agent moves to its desired position. Finally, an attitude controller extracts the desired orientation of each drone and computes the required input torque.

\subsection{Attractive Ellipsoid Method}

The attractive ellipsoid method assures that the system will be bounded by an attractive set which form is defined by a matrix $P$, as is defined by

\begin{definition}
\textbf{Attractive Ellipsoid} \cite{poznyak2014attractive}:		
An ellipsoid, represented by $\varepsilon \subset  \mathbb{R}^n$ with center in $ {\xi}_c$ is given by:
\begin{equation}
	\varepsilon =\left\{  {\xi}\in \mathbb{R}^{n}|\text{ \ }( {\xi}- {\xi}_c)^{\mathsf{T}} {P(\xi-\xi_c)}\leq 1\right\}
\label{ellipsoid}
\end{equation}%
is said to be \textbf{attractive} if for all the trajectories of the vector state $\xi$, it is satisfied that
\begin{equation}
    \limsup_{t\rightarrow \infty } \, { {(\xi-\xi_c)}^{\mathsf{T}} {P(\xi-\xi_c)}\leq 1}
    \label{attractive}
\end{equation}
where the ellipsoidal matrix $ {P}\in \mathbb{R}^{n\times n}$ is a symmetric positive definite matrix $0< P, \,{P=P}^{\mathsf{T} }$.
\end{definition}

If the set \eqref{ellipsoid} is attractive, it is if it satisfies Eq. \eqref{attractive}, the \textbf{practical stability} \cite{lakshmikantham1990practical} of the system is guaranteed, meaning that the system states converge and are bound to a neighborhood around the chosen center of the ellipsoid.

\subsection{Attitude Controller}

Considering the control strategy for the attitude of an aerial vehicle in \cite{looping}, and a desired attitude quaternion $\bar{q}_{id}$, let us define the attitude quaternion error
\begin{equation}
    \bar{q}_{ie} \triangleq \bar{q}_{id}^*\otimes \bar{q}_i= \begin{bmatrix}
        q_{id_0}\,q_{i_0}+q_{id}^\mathrm{T}q_{i} \\
        q_{id_0}\, q_{i}-q_{i_0} q_{id} - q_{id} \times q_{i}
    \end{bmatrix}
    %\begin{bmatrix}
    %    q_{id_0}\,q_{i_0}-q_{id}^\mathrm{T}q_{i} \\
    %    q_{id_0}\, q_{i}+q_{i_0} q_{id} + q_{id} \times q_{i}
    %\end{bmatrix}
\end{equation}
with $\Omega_{ie} \triangleq \Omega_{i}-\Omega_{id}$. From now on, subindex $d$ denotes a desired reference. Using this variables, the error manifolds $s_i$ can be defined as

\begin{equation}
    s_{i} \triangleq \Omega_{ie} + \rho_i q_{ie}
\end{equation}
where $\rho_i$ is a positive diagonal matrix. Then, the attitude controller for each vehicle $\tau_i$ is given by
\begin{equation}
    \tau_i = -K_{di}s_i-\beta_i \mathrm{sat}\left( \gamma_i s_i\right)
    \label{control:attitude}
\end{equation}
where $K_{di}$ is a positive defined matrix, and $\beta_i$ and $\gamma_i$ are positive diagonal matrices.
This controller guarantees the convergence of the error signals to an invariant set \cite{looping}.

%%%DAMR aqui voy
\subsection{Position Controller}
	%mapeo: revisar el cambio de signo del eje Z, hay que cambiar los signos en las ecuaciones
Now, the next step is to estimate the desired attitude signals that will assure the tracking of the position of each vehicle. 
Following the position control strategy for a UAV in \cite{mapeo2021},  let us define a virtual controller $u_{id}=[u_{id1}\ u_{id2}\ u_{id3} ]^T\in \mathbb{R}^3$ for each one of the vehicles dynamics as
	\begin{equation}
		u_{id} \triangleq f_{id} R_{id}(\bar{q}_{id})e_3
		\label{ud}
	\end{equation}
	where the desired thrust $f_{id}$, and the desired attitude quaternion $\bar{q}_{id}$ can be computed as $f_{id} = \Vert u_{id} \Vert $ and
	\begin{equation}
		\overline{q}_{id} =
		\begin{bmatrix}
			{\frac{1}{2} \sqrt{2\hat{u}_{id_3}+2}} \\
			{-\frac{\hat{u}_{id_2}}{ \sqrt{2\hat{u}_{id_3}+2}} } \\
			{\frac{\hat{u}_{id_1}}{ \sqrt{2\hat{u}_{id_3}+2}}} \\ 
			{0}
		\end{bmatrix}
		\label{qdxy}
	\end{equation}
	with $\hat{u}_{id} = u_{id} / \Vert u_{id} \Vert$. Therefore the corresponding desired velocity can be calculated using the time derivative of the desired direction of the thrust $\dot{\hat{u}}_{id}$ as follows
	\begin{equation}
		\Omega_{id} = \footnotesize 
		\begin{bmatrix}
			{{ - \dot{\hat{u}}_{i2} +} \displaystyle\frac{\dot{\hat{u}}_{i3} \hat{u}_{i2} }{\hat{u}_{i3}+1} }\\
			{{  \dot{\hat{u}}_{i1}+} \displaystyle\frac{\dot{\hat{u}}_3  \left(  -\hat{u}_1 \right) }{\hat{u}_{i3}+1}}\\
			{\displaystyle \frac{\hat{u}_{id_1}\dot{\hat{u}}_{id_2}-\hat{u}_{id_2}\dot{\hat{u}}_{id_1}}{\hat{u}_{id_3}+1}}
		\end{bmatrix}
		\label{omegad}
	\end{equation}
	Note that the derivative of the controller $u_{id}$ is needed to compute the desired $\Omega_{id}$.

\subsection{Load Control and Error Dynamics}
Let's define the error variables for the position of the load $x_e$ and the $i$-th vehicle $x_{ei}$ as
\begin{align}
	x_e \triangleq x_L - x_{Ld} \label{error:load}\\
	x_{ei} \triangleq x_i - x_{id}  \label{error:quadri}
\end{align}
from the system constrains \eqref{model:restriction}, the desired position of each vehicle is
\begin{equation}
	x_{id} \triangleq x_{Ld}-L_i \alpha_{id}.
	\label{xid}
\end{equation}
using the 2nd time derivative of \eqref{error:load}, and substituting \eqref{model:load} we obtain
\begin{align}
	\ddot{x}_e &= \ddot{x}_L - \ddot{x}_{Ld}\\
	\ddot{x}_e &= -  \frac{1}{m_L}  \sum_{i=1}^{n}  \left( T_i\alpha_i \right) - g\,e_3 -\ddot{x}_{Ld} \label{error:load2}
\end{align}
now, consider a virtual controller for the load defined as
\begin{equation}
	u_L \triangleq  - \sum_{i=1}^{n}  \left( T_{id} \alpha_{id} \right)
	\label{virtual:load}
\end{equation}
adding and subtracting \eqref{virtual:load} in \eqref{error:load2} results in the open-loop error dynamic of the load
\begin{equation}
	\ddot{x}_e = -  \frac{1}{m_L} \left( u_L - \sum_{i=1}^{n}  \left( \zeta_{Li} \right) \right)  - g\,e_3 -\ddot{x}_{Ld}
\end{equation}
%from the quadrotor dynamics \eqref{model:pos}, isolating $T_i\alpha_i$ and $\ddot{x}_{ei}=\ddot{x}_i-\ddot{x}_{id}=\ddot{x}_i-\ddot{x}_{Ld}+L_i\ddot{\alpha}_{id}$ the open-loop error dynamics is
%
%\begin{align}
%	\ddot{x}_e &=- \frac{1}{m_L} \left( u_L - \sum_{i=1}^{n}  \left( m_i\ddot{x}_i -f_iR_ie_3+ m_i g e_3 -  T_{id}\alpha_{id} \right) \right)  - g\,e_3 -\ddot{x}_{Ld}\\
%	\begin{split}
%		\ddot{x}_e &= - \frac{1}{m_L} \left( u_L - \sum_{i=1}^{n}  \left( m_i \ddot{x}_{ei}+m_i\ddot{x}_{Ld} -m_iL_i\ddot{\alpha}_{id}-u_i - \zeta_i + m_i g e_3 - T_{id}\alpha_{id} \right) \right) \\
%		&- g\,e_3 -\ddot{x}_{Ld} 
%	\end{split}
%\end{align}
%
where $\zeta_{Li}$ is the error between the actual and desired tension defined as
\begin{equation}
\zeta_{Li} \triangleq T_i\alpha_i -  T_{id}\alpha_{id}
\end{equation}

Note that the disturbance on the load corresponds to the sum of the error in the tensions on each vehicle. 

Now, for the error dynamics in each vehicle, let us consider the second time derivative of the error position from each UAV dynamics \eqref{error:quadri} and \eqref{xid}, also, adding and subtracting the virtual position controller \eqref{ud} results in
\begin{equation}
	\ddot{x}_{ei} = \frac{1}{m_i}u_i +\frac{1}{m_i}\zeta_i-ge_3 +\frac{1}{m_i}T_i\alpha_i - \ddot{x}_{Ld} + L_i\ddot{\alpha}_{id}
\end{equation}
where
\begin{equation}
	\zeta_i \triangleq f_iR_i e_3 - f_{id}R_{id}e_3
\end{equation}
Note that the disturbance $\zeta_i$ is the difference between the desired and the actual input force on each vehicle. Assuming that the desired thrust $f_{id}$ is commanded instantaneously, then, $\zeta_i$ depends on the attitude tracking error. Therefore, as $\bar{q}_{ie} \rightarrow [1\ 0\ 0\ 0]^T$ this disturbance $\zeta_i \rightarrow 0$ \cite{mapeo2021}.

\subsection{Closed-loop Error Dynamics}

Let us propose the control laws for the virtual controllers $u_L$ and $u_i$ as
\begin{equation}
	\begin{split}
		u_L=-m_L\left( ge_3+\ddot x_{Ld}\right)-\nu_L
	\end{split}
	\label{control:uL}
\end{equation}
\begin{equation}
	\begin{split}
		u_i=m_i \left( ge_3 + \ddot x_{Ld} \right) - T_{id}\alpha_{id} + \nu_i
	\end{split}
	\label{control:ui}
\end{equation}
where $$\nu_L \triangleq -k_{pL}(x_L-x_{Ld}) -k_{dL}(\dot x_L-\dot x_{Ld})-k_{iL}\int (x_L-x_{Ld})$$ and $$\nu_i \triangleq  -k_{pi}(x_i-x_{Ld}+L_i\alpha_{id}) -k_{di}(\dot x_i-\dot x_{Ld}+L_i\dot{\alpha}_{id}) - k_{ii}\int (x_i-x_{Ld}+L_i\alpha_{id}),$$ with the control gains  matrices $k_{pL},k_{dL},k_{iL},k_{pi},k_{di},k_{ii}>0\in\mathbb{R}^{3 \times 3}$. Please note that the desired UAVs positions are given by the desired load's position $x_{Ld}$ and the desired cable orientation $\alpha_{id}$. Then, the closed-loop error dynamics of the systems result in
%Assuming that $\ddot{\alpha}_{id} \approx 0$

%\begin{align}
%	\ddot{x}_e &= - \frac{1}{m_L} \left(-\sum_{i=1}^{n} \left( \zeta_{Li}- \zeta_i + \nu_i(x_{ei} )\right) + \nu_L(x_e)\right) 
%\end{align}

\begin{align}
	\ddot{x}_{e} &= \frac{1}{m_L} \left( \sum_{i=1}^{n}  \left( \zeta_{Li} \right)+ \nu_L \right)
\end{align}

\begin{align}
	\ddot{x}_{ei} &= \frac{1}{m_i} \left(\nu_i +\zeta_i  +\zeta_{Li} \right)+L_i\ddot{\alpha}_{id}
\end{align}
%where the new control law takes the form $\nu_i \triangleq k_{pi} e_{ei} + k_{di} \dot e_{ei} + k_{ii} \int  e_{ei}$, and $\nu_L \triangleq k_{pL} e_{e} + k_{dL} \dot e_{e} + k_{iL} \int  e_{e}$, with $k_{pi},k_{di},k_{ii},k_{pL},k_{dL},k_{iL}\in \mathbb{R}$ as control gains.
%Also, $d_{i} \triangleq  T_i\alpha_i -  T_{id}\alpha_{id} $.
Now let's define a state space based on the error dynamics as

\begin{equation}
	\chi \triangleq \left[\int x_e ^{\mathrm{T}} dt, x_e ^{\mathrm{T}}, \dot{x}_e ^{\mathrm{T}}, \int x_{e1} ^{\mathrm{T}} dt, x_{e1}^{\mathrm{T}}, \dot{x}_{e1}^{\mathrm{T}},\hdots,\int x_{en}^{\mathrm{T}} dt, x_{en}^{\mathrm{T}},\dot{x}_{en}^{\mathrm{T}}\right]^{\mathrm{T}}\in \mathbb{R}^{9(n+1)}
\end{equation}
and its time derivative
\begin{equation} \dot{\chi} =
	\tilde{A} \chi + \tilde{B} \zeta
	\label{model:errorfullstate}
\end{equation}
where
\begin{equation*}
	%	\begin{split}
	\tilde{A} \triangleq \begin{bmatrix}
		A+\frac{1}{m_L}K_L & 0 & \hdots & 0\\
		0 & A+\frac{1}{m_1}K_1 & \hdots & 0 \\
		\vdots & \vdots & \ddots & \vdots \\
		0 & 0 & \hdots & A+\frac{1}{m_n}K_n
	\end{bmatrix} , \end{equation*}
\begin{equation*}
	\tilde{B} \triangleq \begin{bmatrix}
	\frac{1}{m_L}B & 0 & 0 & \frac{1}{m_L}B & 0 & 0 &\hdots & \frac{1}{m_L}B & 0 & 0\\
	\frac{1}{m_1}B & \frac{1}{m_1}B & L_1B & 0 & 0 & 0 &\hdots & 0 & 0 & 0\\
	0 & 0 & 0 & \frac{1}{m_2}B & \frac{1}{m_2}B & L_2B &\hdots & 0 & 0 & 0\\
	\vdots & \vdots & \vdots & \vdots & \vdots& \vdots & \ddots & \vdots & \vdots & \vdots  \\
	0 & 0 & 0 & 0 & 0 & 0&\hdots & \frac{1}{m_n}B & \frac{1}{m_n}B & L_nB
\end{bmatrix}, \zeta \triangleq \begin{bmatrix}
\zeta_{L1} \\
\zeta_1\\
\ddot{\alpha}_{1d} \\
\vdots \\
\zeta_{Ln} \\
\zeta_n\\
\ddot{\alpha}_{nd}
\end{bmatrix}\in \mathbb{R}^{9n} , \end{equation*} \begin{equation*}
	A \triangleq \begin{bmatrix}
		0 & I & 0 \\
		0 & 0 & I\\
		0 & 0 & 0
	\end{bmatrix} , 
	B \triangleq \begin{bmatrix}
	0  \\
	0 \\
	I 
\end{bmatrix} , 
	K_L \triangleq \begin{bmatrix}
	0 & 0 & 0\\
	0 & 0 & 0\\
	-k_{iL} & -k_{pL} & -k_{dL}
\end{bmatrix} ,
	K_i \triangleq \begin{bmatrix}
	0 & 0 & 0 \\
	0 & 0 & 0\\
	-k_{ii} & -k_{pi} & -k_{di}
\end{bmatrix} 
%	\end{split}
\end{equation*}
with the identity matrix $I \in \mathbb{R}^{3 \times 3}$.

%In the control strategy the desired $\alpha_{id}$ depends on the load control $u_L$, the following assumptions are considered
%\begin{itemize}
%	\item Small initial errors imply small changes in the controller, then the 2nd derivative of the PID control is bounded.
%	\item The desired acceleration $\ddot{x}_{Ld}$ of the load is bounded.
%	\item Then the control $u_L$ in \eqref{control:uL} is bounded, and also its 2n derivative.
%	\item Therefore there exist a constant that represents the bound of the 2nd derivative of $\ddot{\alpha}_{id}$
%	\begin{equation}
%		\Vert \ddot{\alpha}_{id} \Vert < \alpha_{id}^+
%	\end{equation}
%\end{itemize}
\subsubsection{Assumptions}
\begin{enumerate}
%	\item 
	\item The attitude controllers of each UAV \eqref{control:attitude} assure the errors to be attracted and confined into a small invariant set \cite{looping}. Therefore, the position virtual controller is induced, with a small disturbance $\zeta_i$, with a bound $c_{1i}$, it is
\begin{equation}
	\Vert \zeta_i \Vert ^2 \leq c_{1i}
 \end{equation}
	\item The desired position of the vehicles $x_{id}$ is calculated using $\alpha_{id}$ which is estimated from the payload controller $u_L$. Also, considering well-posed initial conditions in compliance with the constrains \eqref{model:restriction}, along with small initial errors, and assuming that the desired trajectory of the load ${x}_{Ld}$ is smooth and slow varying, then it is assumed that the controller of the load have a bounded transient response, implying that the second time derivative of $\alpha_{id}$ is bounded
\begin{equation}
	\Vert \ddot{\alpha}_{id} \Vert ^2 \leq c_{2i}
\end{equation}
	\item The difference between the desired and actual tensions on the cables are bounded
	\begin{equation}
		\Vert \zeta_{Li} \Vert ^2 \leq c_{3i}
	\end{equation}
\end{enumerate}
with $\{c_{1i},c_{2i},c_{3i}\}>0 \in \mathbb{R}$ as constant positive bounds.
\subsection{Stability Analysis}
\label{sec:stability}

\begin{theorem}
		For the system \eqref{model:pos} and \eqref{model:load} with the restriction \eqref{model:restriction} and the virtual controls \eqref{control:uL} and \eqref{control:ui}, under assumptions $1-3$, and considering some matrices $\tilde{A}$ and $\tilde{B}$, if there exists a symmetric positive definite matrix $P \in \mathbb{R}^{9(n+1) \times 9(n+1)}$, $P=P^{\mathsf{T}}>0$, and some positive constants $\alpha$ and $\varepsilon$ that satisfy the inequality 
  $$W_L(P,K_L, K_1,...,K_n,\alpha, \varepsilon)<0$$
		where
		\begin{equation}
			W_L=\begin{bmatrix}
				{P\tilde{A}+\tilde{A}^{\mathsf{T}}P+\alpha P}&{P\tilde{B}}\\{\tilde{B}^{\mathsf{T}}P}&{-\varepsilon I_{9n\times 9n}}
			\end{bmatrix},
		\end{equation}
	 then, there exists an attractive stability region around the origin of the vector state $\chi$ defined by the ellipsoid
	\begin{equation}
		\mathcal{E} =\left\{ \chi \in \mathbb{R}^{n}| \; \chi^{\mathsf{T}}P \chi \leq \frac{\beta}{\alpha} \right\}
	\end{equation}
		where $\beta>0$. Henceforth, the practical stability of the system is guaranteed.
\end{theorem}

\begin{proof}
	In order to analyze the system, the following Lyapunov-like energy function is proposed
	\begin{equation}
		V= \chi ^{\mathsf{T}} P \chi
	\end{equation}
	whose derivative, 
	%using \eqref{cotaPerturbacionPos}, 
	adding and subtracting $\alpha V$ and $\varepsilon \Vert {\zeta}\Vert^2$, using assumptions 1-3 
	%for the bound of the disturbances $\varepsilon \Vert {\zeta}_i \Vert^2 \leq \varepsilon_{\zeta_i} \Vert {\zeta}_i \Vert^2+\varepsilon d^{+} +\varepsilon \eta^{+}$, 
	is equal to
	\begin{equation}
		\begin{array}{rcl}
			\dot{V} &=& \chi ^{\mathsf{T}} P \dot{\chi} +\dot{\chi} ^{\mathsf{T}} P \chi \pm \alpha V \pm \varepsilon \Vert {\zeta} \Vert^2
			\\ &\leq& \begin{bmatrix}
				{\chi}\\{{\zeta}}
			\end{bmatrix}^{\mathsf{T}}
			W_L
			\begin{bmatrix}
				{\chi}\\{{\zeta}}
			\end{bmatrix}- \alpha V + \beta 
		\end{array}
	\end{equation}
	where $\beta = \varepsilon \sum_{i=1}^{n} \left( c_{1i} + c_{2i}+c_{3i}\right)$.
	As long as it is assured that $W_L \leq 0$, there exists an attractive stability region around the origin of $x_e$ and ${x}_{ei}$ defined by the ellipsoid
	%Mientras se asegure que la matriz $W_p \leq 0$, se confirma que existe una región de estabilidad alrededor del origen de $x_p$ y $\hat{x}_e$ definida por el elipsoide %de  $\alpha$, $\beta$ y $P$.
	\begin{equation}
		\mathcal{E} =\left\{ \chi \in \mathbb{R}^{9(n+1)}| \; \chi ^{\mathsf{T}}P \chi \leq \frac{\beta}{\alpha} \right\}
	\end{equation}%
	and the solution to the optimization problem is 
	%por lo que, la solución al problema de optimización
	%
	\begin{equation}
		\begin{array}{c}
			\mathrm{tr}\left\lbrace \frac{\beta }{\alpha }P^{-1}\right\} \rightarrow
			{\min\limits_{\alpha ,\beta ,\varepsilon, K_L, K_1,...,K_n, P} } \\
			\text{{\small subject to the restrictions}} \\
			\alpha >0,\text{ }\varepsilon>0,\\
			0<P,\text{ }W_L=W_L\left(  K_L, K_1,...,K_n,P,\alpha ,\varepsilon, \right)
			\leq 0%
		\end{array}
		\label{OP-Load}
	\end{equation}%
Therefore, if there exist the solution to the problem \eqref{OP-Load} the practical stability is guaranteed that correspond to the smallest attractive region.

\end{proof}

% Discusion de la prueba de estabilidad
Note the vector $\zeta$ is composed with all the disturbances present in the closed-loop system. $\zeta_{Li}$ corresponds to the difference between the actual and the desired tension in the cable, and appears in the error dynamics of the load and each vehicle, this tension increases as the error in position $x_{ei}$ increases, meaning that as long as the position controller assures small errors this disturbance would be small enough to assure practical stability.

$\ddot{\alpha}_{id}$ corresponds to the variation in the direction of the desired tension and only appears in the vehicles error dynamics. This implies that the transient response of the control load directly affects the bound of this disturbance. Therefore, non continuous desired position trajectories should not be used, also, the load virtual controller should be as damped as possible, this results in reducing the upper bound of the $\ddot{\alpha}_{id}$. 

$\zeta_i$ is the disturbance that represents the difference of the desired and actual input forces in each vehicle. 
Given that the attitude dynamics of the vehicles are faster than the position dynamics, initial small errors in the attitude controller are desirable, in order to assure that this disturbance is small enough and the position will converge to the desired one, even if there are small disturbances of $\zeta_{Li}$ and $\ddot{\alpha}_{id}$. 
The attitude controller has to be highly reactive in order to assure the tracking of the desired trajectory, however, if the position controller is highly reactive, it will induce aggressive attitude trajectories that will result in greater attitude errors. 
Therefore there exists a compromise between the controllers, as the attitude controller must be highly reactive and assure the tracking for different trajectories, then the position controller must be reactive enough to assure the convergence of the position compensating the disturbances with smooth transient response, and finally the load controller must have a damped response.

\begin{table}[b!]
\caption{Simulation parameters in International System units. Index $i\in\{1,2,3\}$. Function $diag(*)$ is a square diagonal matrix whose diagonal values are given by the argument, and $\beta_i =  \mathbf{0}$.}
 \begin{center}
\begin{tabular}{  c  c  c  c }
 \hline                 
  $m_i $ & $m_L$ & $J_i $ &  $L_i$\\
 \hline  
   $0.5$ & $0.225$ & $0.01 diag(2.32,2.32,4)$ & $1$ \\
  \hline \hline
  $k_{pi} $ & $k_{di}$ & $k_{ii}$ & $\rho_{i}$  \\
  \hline
 $40[1\ 1\ 1.5]^T$ & $10[1\ 1\ 1.2]^T$ & $2[1\ 1\ 2]^T$ &  $62.5I$ \\
  \hline\hline
  $k_{pL} $ & $k_{dL}$ & $k_{iL}$ & $K_{di}$ \\  
    \hline
 $9[1\ 1\ 1]^T$ & $3.5[1\ 1\ 1]^T$ & $0.2[1\ 1\ 1]^T$ & $16I$\\
  \hline
 \end{tabular}
\end{center}
\label{tabla:param}
\end{table}

\section{Numerical Simulations}
\label{sec:simulation}
\begin{figure}[h!]
    \centerline{\includegraphics[width=0.98\textwidth, angle=90]{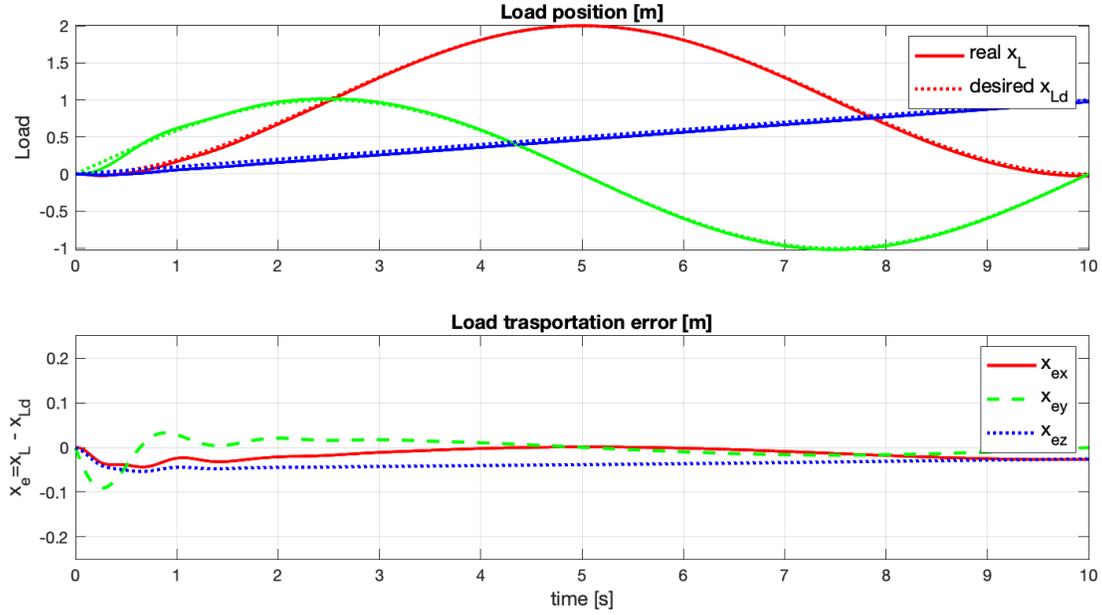}}
    \caption{Load position $x_L$ (top) and load transportation error $x_e=x_L-x_{Ld}$ (bottom). }
    \label{fig:load_pos}
\end{figure}
\begin{figure}[h!]
    \centerline{\includegraphics[width=0.98\textwidth, angle=90]{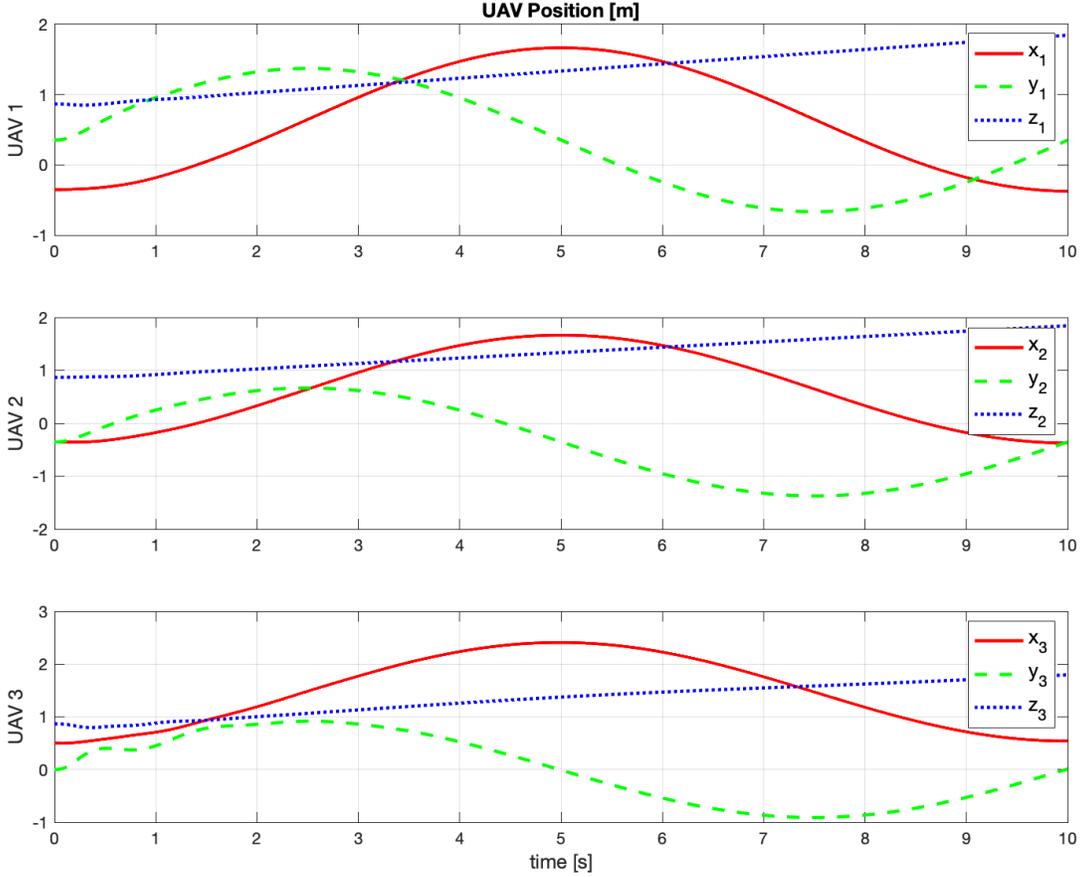}}
    \caption{UAV's positions $x_i$.}
    \label{fig:UAV_pos}
\end{figure}
\begin{figure}[h!]
    \centerline{\includegraphics[width=0.98\textwidth, angle=90]{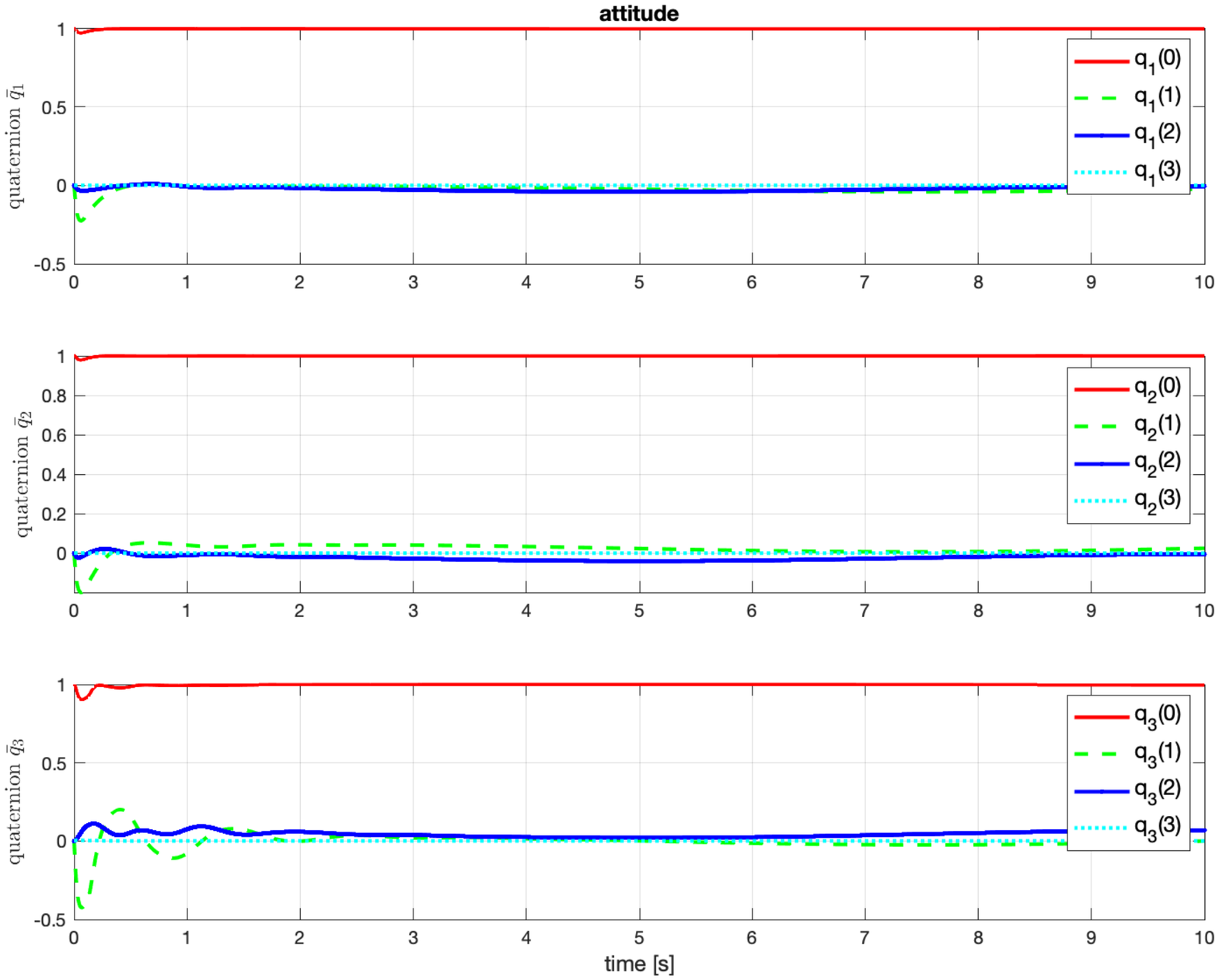}}
    \caption{UAVs' attitude quaternion $\bar q_i$.}
    \label{fig:load_quat}
\end{figure}
In order to validate the control strategy proposed in Section \ref{sec:control}, and to study the behavior of the closed-loop system, numerical simulations were carried out with the help of Matlab/Simulink\textregistered. A video showing the simulation is available at \url{https://youtu.be/uzAYqm1-H_U}. As a case of study, let us consider three UAVs carrying the payload. The main parameters used in the simulation are presented in Table \ref{tabla:param}.
\begin{figure}[h!]
    \centerline{\includegraphics[width=0.98\textwidth, angle=90]{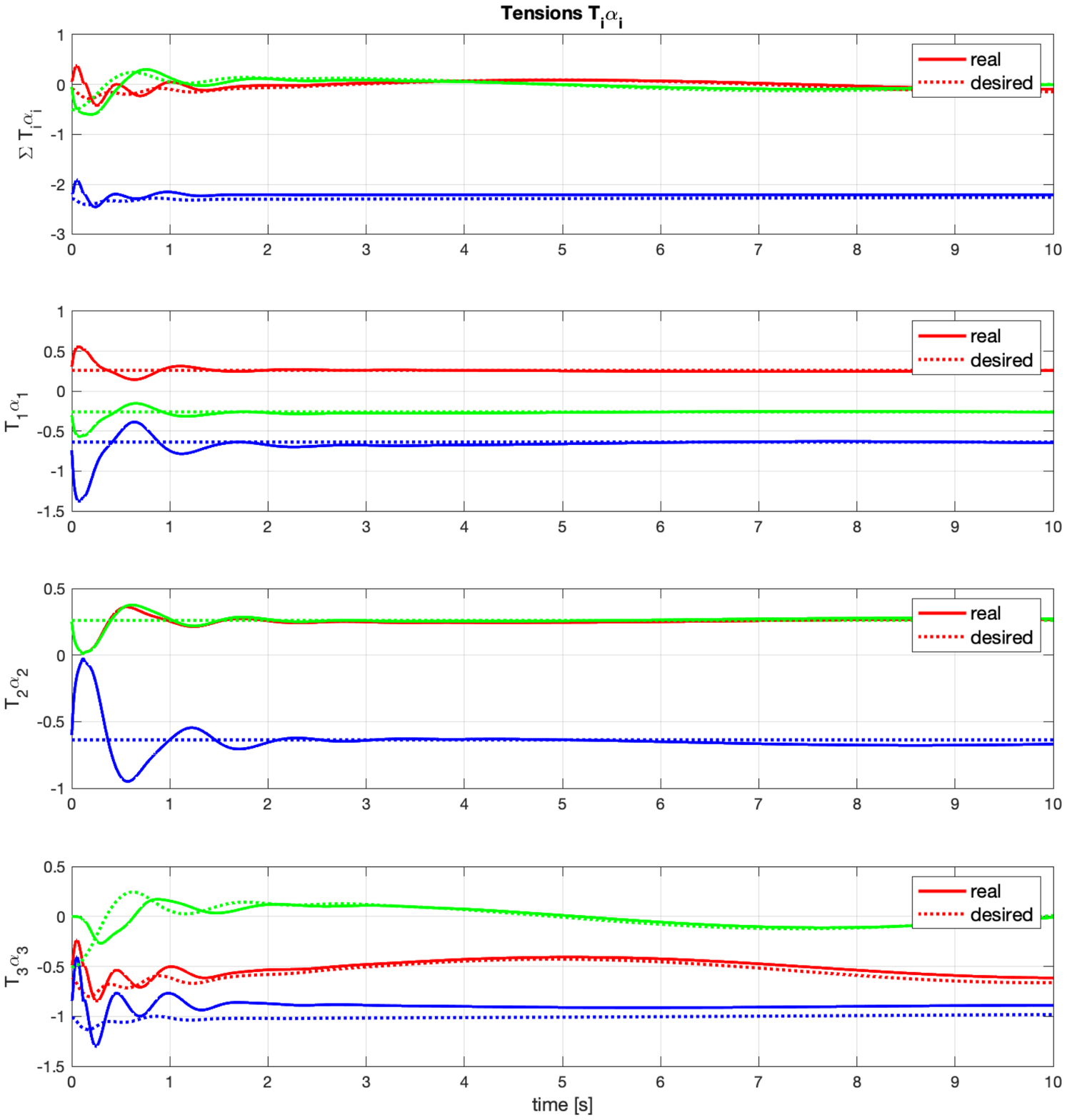}}
    \caption{Tensions $T_i\alpha_i$. The solid lines represent the actual signals, while the dashed lines represent the desired references in the three axes.}
    \label{fig:load_Tqi}
\end{figure}

The objective is for the load to track a desired trajectory, in this case an ascending spiral (see Fig. \ref{fig:load_pos}) of the form

\begin{equation}
x_{Ld}= \begin{bmatrix} 1- \cos(\frac{2}{5}\pi t )\\
                  \sin (  \frac{2}{5}\pi t )\\
                   t /10
    \end{bmatrix}
    \label{eq:traj}
\end{equation}
where $t$ is the time variable. The initial positions of the UAVs and the load are
\begin{eqnarray}
        x_L=[0\ 0\ 0]^T\\
        x_1= R_{(z,-\pi/4)}R_{(y,-\pi/6)} [0\ 0\ L_1]^T\\
        x_2= R_{(z,\pi/4)}R_{(y,-\pi/6)} [0\ 0\ L_2]^T\\
        x_3= R_{(y,\pi/6)} [0\ 0\ L_3]^T
\end{eqnarray}
with $R_{(a,b)}\in SO(3)$ as the basic rotation matrices representing a rotation of an angle $b\in \mathbb{S}^1$ around one of the main axis $a\in \{x,y,z\}$. Note that from Eqs. (\ref{virtual:load}), (\ref{control:uL}) we can obtain the desired tension on the cables, but this results in an under-constrained system with infinite solutions, then we can impose additional constrains to remain with an unique solution, in this case, the desired tensions are selected as follows
\begin{eqnarray}
        T_1\alpha_{1d}=- \frac{m_L}{3}R_{(z,-\pi/4)}R_{(y,-\pi/6)} (ge_3) \\\label{eq:t1}
        T_2\alpha_{2d}=- \frac{m_L}{3}R_{(z,\pi/4)}R_{(y,-\pi/6)} (ge_3)
        \\\label{eq:t2}
        T_3\alpha_{3d}=-u_L-T_1\alpha_{1d}-T_2\alpha_{2d}
        \label{eq:t3}
\end{eqnarray}
% \begin{figure}[t]
%     \centerline{\includegraphics[width=0.98\textwidth, angle=90]{imgs/oriLoad.pdf}}
%     \caption{UAVs' attitude in Euler angles $\phi,\theta,\psi$.}
%     \label{fig:load_ori}
% \end{figure}
\begin{figure}[h!]    \centerline{\includegraphics[width=0.98\textwidth, angle=90]{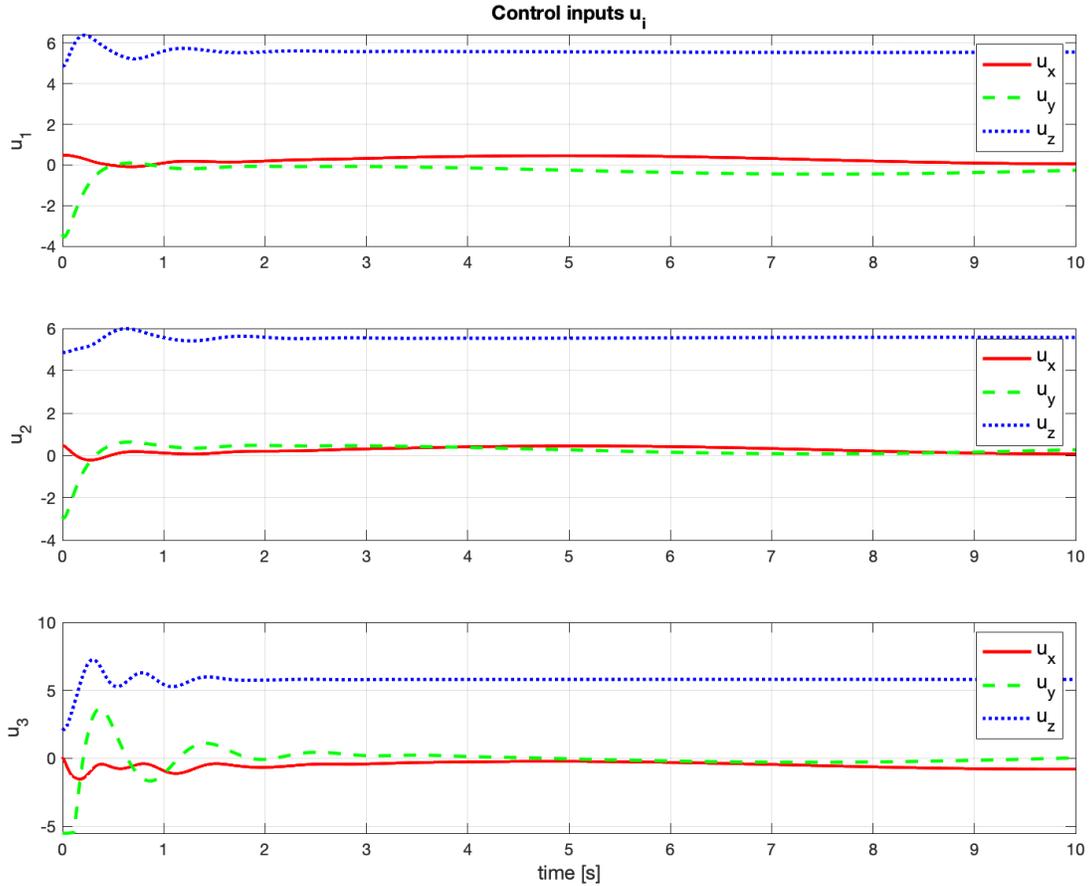}}
    \caption{Control inputs $u_i$.}
    \label{fig:load_u}
\end{figure}
any other valid solution can be used instead, for instance, it would be interesting to select the solution that minimizes the total energy among the $n$ agents. 

\begin{figure}[h!]
    \centerline{\includegraphics[width=0.98\textwidth, angle=90]{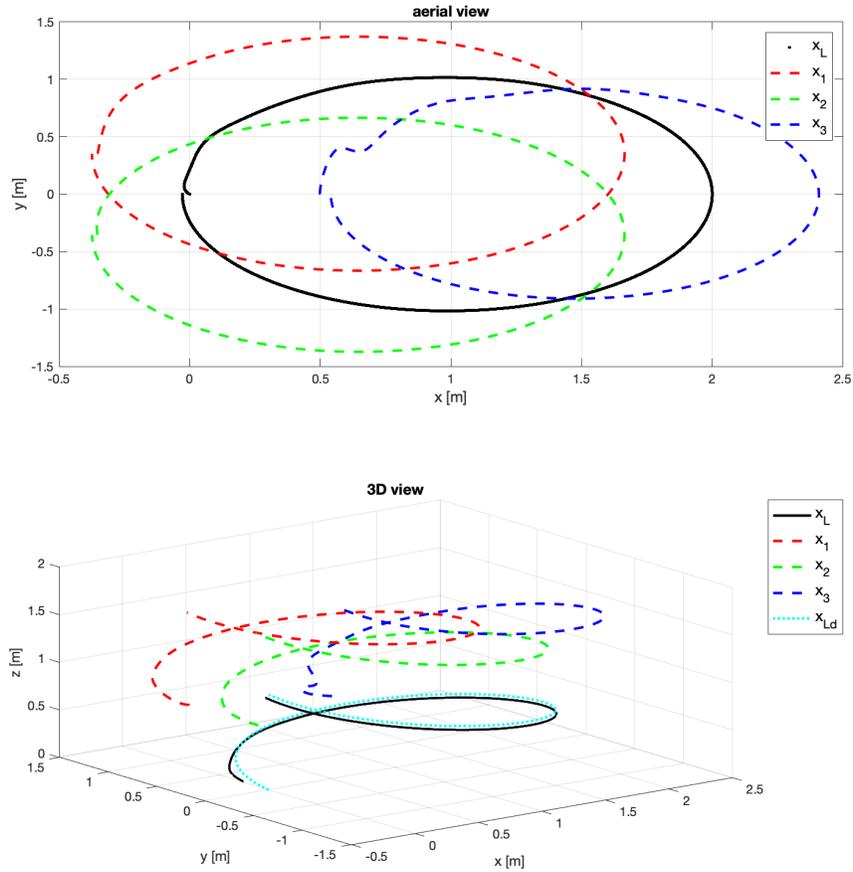}}
    \caption{Trajectories of the cable suspended transportation system with 3 UAVs. Aerial view ($x-y$ plane) on top, 3D view at the bottom. The control strategy demonstrated good performance, being able to accurately track the desired load's trajectory.}
    \label{fig:load_3d}
\end{figure}

Figures \ref{fig:load_pos}-\ref{fig:load_3d} demonstrate the good performance of the control strategy. At the top of Fig. \ref{fig:load_pos} the load position is presented, where we can appreciate how the load follows the desired reference, while the payload error $x_e$ is depicted at the bottom. We can observe that the errors converge to the invariant set, hence, they remain small and bounded, which is in compliance with practical stability.

The UAVs' positions along the mission are shown in Fig. \ref{fig:UAV_pos}, where we can see how the third UAV has to correct its initial formation error. Also,  in Fig. \ref{fig:load_quat} the UAVs' attitude quaternions are presented. Moreover, in Fig. \ref{fig:load_Tqi} the desired (dotted lines) and real tensions (solid lines), as well as the total tension (top) are depicted. We can note that the tension signals are smooth and remain bounded. 
%\textcolor{blue}{XXXXXX aqui decir tambien que se cumple la suposicion sobre la aceleracion en los cables XXXXXX}

The drones' control inputs are shown in Fig. \ref{fig:load_u}. Here we can observe that due to the choice of the desired tensions in Eqs. \eqref{eq:t1}-\eqref{eq:t3}, the third UAV must make a bigger control effort to compensate for the errors in the load positioning, besides its larger initial error. Nevertheless, all the UAVs converge quickly to their desired trajectories, and remain within small bounded errors.

Finally, we can appreciate the good overall transportation system performance in Fig. \ref{fig:load_3d}, where the top view (top) and the 3D view (bottom) of the UAVs and load trajectories are presented. 

%\textcolor{blue}{XXXXX falta un parrafo concluyendo las simulaciones XXXX}
The numerical results in this case of study support the validity of the approach, and that the closed-loop coupled system is stable and the position of the load converges to the desired. Also, as discussed in Section \ref{sec:stability}, the system disturbances are larger in magnitude at the beginning, during the transient state. Even in this scenario, the error signals remain small and bounded, corroborating practical stability of the system.
%
%-control effort in uav3
%-small and bounded postion errors
%-smooth trajectory

\section{Conclusions}
\label{sec:conclusions}

In this work a hierarchical controller for the transportation of a cable suspended load by multiple UAVs is presented.  
The proposed algorithm considers the collaboration of $n$ vehicles, using a continuous controller that assures that the error signals are confined to an attractive invariant set, assuring practical stability for the coupled system, resulting in the tracking of a time varying trajectory by the load. To do so, the desired sum of the cable tensions is used as a virtual controller for the load trajectory tracking, then, the desired tensions and direction for each cable can be obtained, resulting in an under-determined system, where additional constrains can be added to obtain an unique solution, thereafter, from the cables' direction we compute the desired position for each UAV agent, which is in turn fed to a position controller using the desired thrust vector as a new virtual controller, from which the desired attitude of each drone are drawn and parsed to a quaternion based attitude controller. 

The proposed algorithm, based in the attractive ellipsoid method, considers two disturbances due to the virtual controllers, one that depends on the attitude error, and the other one that depends on the position error. By means of a Lyapunov-like energy function, we have studied the stability of the closed-loop system, demonstrating its practical stability, and establishing some sufficient conditions in order to guarantee small bounds in the errors.

Numerical simulation were presented in order to demonstrate the viability of the proposed scheme, and study its performance, showing the tracking of a ascending spiral trajectory by the load carried by three UAVs, obtaining good performance with small bounded errors. From the stability analysis and the simulations study, we have encounter that the moment when the disturbances are more significant is during their initial state, given that both the attitude and position controllers had not converged jet to their invariant sets during the transient state, thus, small initial errors are required in order to assure that this disturbances are bounded and small.

As a future work, we are working on the validation of the proposed control strategy in real-time experiments with three or more drones transporting a load. Also, it would be interesting to propose other suitable approaches to obtain the desired cable tensions from the resultant tension, for instance to minimize the total energy. Finally, we would like to study different control approaches applied to this kind of multi-agent transportation system.

 %\appendix

\section{Declaration of Competing Interest}
The authors declare that they have no known competing financial interests or personal
relationships that could have appeared to influence the work reported in this paper.

  \bibliographystyle{elsarticle-num} 
  \bibliography{suspendedLoad}





%\end{thebibliography}
\end{document}